\documentclass[12pt]{Epigenetics} 

\usepackage[english]{babel} 
\usepackage{graphicx}

\definecolor{color1}{RGB}{0,0,90} 
\definecolor{color2}{RGB}{119, 136, 153} 

\usepackage{hyperref} 
\hypersetup{hidelinks,colorlinks,breaklinks=true,urlcolor=color1,citecolor=color1,linkcolor=color1,bookmarksopen=false,pdftitle={Title},pdfauthor={Author}}

\usepackage[square,numbers]{natbib}
\usepackage{appendix}

\DeclareMathOperator*{\argmin}{arg\,min}

\RequirePackage{algorithm} 
\RequirePackage{algorithmic} 
\RequirePackage{multirow}
\RequirePackage{rotating}
\RequirePackage{mathtools, nccmath}
\RequirePackage{float}
\RequirePackage{symbol}
\RequirePackage[symbol]{footmisc}

\newtheorem{theorem}{Theorem}

\def \applip 		{Appendix A}
\def \apptheorem 	{Appendix B}
\def \appnorm 		{Appendix C}
\def \applambda 		{Appendix D}
\def \appgamma 		{Appendix E}

\PaperTitle{An Efficient Smoothing Proximal Gradient Algorithm for Convex Clustering } 

\Authors{\stepcounter{footnote}Xin Zhou\thanks{Department of Electrical and Computer Engineering, University of Miami, Coral Gables, Florida 33146, United States}\,, \stepcounter{footnote}Chunlei Du\footnotemark[2]\,, and \stepcounter{footnote}Xiaodong Cai\footnote[1]{x.cai@miami.edu}\textsuperscript{\,\,,}\footnotemark[2]} 

\newcommand{\keywordname}{Keywords} 

\Abstract{Cluster analysis organizes data into sensible groupings and is one of fundamental modes of understanding and learning.  
The widely used K-means and hierarchical clustering methods can be dramatically suboptimal due to local minima. Recently introduced convex clustering approach formulates clustering as a convex optimization problem and ensures a globally optimal solution. However, the state-of-the-art convex clustering algorithms,  based on the alternating direction method of multipliers (ADMM) or the alternating minimization algorithm (AMA), require large computation and memory space, which limits their applications. In this paper, we develop a very efficient smoothing proximal gradient algorithm (Sproga) for convex clustering. Our Sproga is faster than ADMM- or AMA-based convex clustering algorithms by one to two orders of magnitude. The memory space required  by Sproga is less than that required by ADMM and AMA  by at least one order of magnitude.  Computer simulations and real data analysis show that Sproga outperforms several well known clustering algorithms including K-means and hierarchical clustering.   The efficiency and superior performance of our algorithm will help convex clustering to find its wide application. }

\begin{document}

\flushbottom 
\maketitle 

\thispagestyle{empty} 
\section{Introduction}
Human beings are skilled at dividing objects into groups or clusters based on certain characteristics of the objects. For example, even children can quickly assign the objects in pictures to groups such as vehicle, animal, people, and building, etc.  Grouping or clustering is therefore  not only helpful to the organization of objects in a meaningful manner, but also fundamental to the understanding and learning of the natural structure of objects. Similarly, data clustering is a fundamental problem in machine learning. The aim of data clustering is to partition a given set of data points into subsets or clusters, such that those within each cluster are more similar to one another than those  assigned to different clusters.  Many clustering methods have been developed \cite{hennig2015handbook, aggarwal2014data}, including the K-means \cite{macqueen1967some}, hierarchical clustering \cite{johnson1967hierarchical}, the density-based method such as DBSCAN \cite{ester1996density}, and spectral clustering \cite{von2007tutorial}, to name a few.  However, the widely used K-means and hierarchical clustering methods are not robust, since their result depends on random initialization and can be dramatically suboptimal due to local minima. Recent studies \cite{krzak2019benchmark,kou2014evaluation,pirim2012clustering} based on simulated and real data showed that no single clustering algorithm can consistently outperform other clustering algorithms on all different types of data. 

Recently, a new approach to clustering named convex clustering has been proposed \cite{hocking2011clusterpath,lindsten2011just, chi2015splitting}. It formulates the clustering problem as a convex optimization problem, thereby guaranteeing to find the optimal solution \cite{tan2015statistical}. Convex clustering can provide an entire clustering path that can be visualized with a dendrogram \cite{weylandt2019dynamic}, similar to what the agglomerative hierarchical clustering offers, but with better computational efficiency and global optimality. Moreover, it can perform clustering and feature selection simultaneously \cite{wang2018sparse}, which may help to select appropriate features to improve clustering performance. Despite these attractive features, convex clustering has not been used widely, because the state-of-the-art convex clustering algorithms based on the alternating direction method of multipliers (ADMM) or the alternating minimization algorithm (AMA) \cite{chi2015splitting,wang2018sparse} demand large computation and memory, especially when the number of data samples $n$ and the dimension of the data points $p$ are large.

In this paper, we employ the smoothing technique for non-smooth convex optimization \cite{nesterov2005smooth} to develop a  novel convex clustering algorithm,  named smoothing proximal gradient algorithm (Sproga), which is much faster and requires much less memory comparing with the convex clustering algorihtms based on ADMM and AMA \cite{chi2015splitting,wang2018sparse}. Similar to the S-ADMM and S-AMA algorithms \cite{wang2018sparse}, our Sproga algorithm can conduct clustering and feature selection simultaneously.
Our computer simulations and real data analysis demonstrate that Sproga outperforms several popular clustering algorithms including the K-means, hierachical clustering, DBSCAN \cite{ester1996density}, a spectral clustering algorithm named SPECC \cite{ng2002spectral}, and a graph based clustering algorithm named Louvain \cite{blondel2008fast}.

\section{Methods}
\subsection{Convex Clustering Formulation}
Suppose that we have $n$ data points $\bbx_1, \bbx_2, ..., \bbx_n \in \mathbb{R}^p$ to be clustered. The  convex clustering approach is formulated as the following convex optimization problem \cite{hocking2011clusterpath,lindsten2011just, chi2015splitting}:
\begin{equation}
\label{eq:cc}
\begin{split}
    \min_{\bbU \in \mathbb{R}^{p \times n}} \frac{1}{2}\sum_{i=1}^n \lVert\bbx_i - \bbu_i \rVert_2^2 + \lambda \sum_{i < j}\omega_{ij}\lVert\bbu_i - \bbu_j\rVert_q ,
\end{split}
\end{equation}
where $\bbu_i\in \mathbb{R}^p$ is the cluster center attached to $\bbx_i$, $\bbU = [\bbu_1, \bbu_2, ..., \bbu_n]$, $\lambda >0$ is a tuning parameter, $\omega_{i,j}$'s are constants
that will be specified later, and $\lVert \cdot \rVert_q$ denotes the $q$-norm with $q=1, 2, \cdots, \infty$.  The second term in the objective function is similar to the fused lasso penalty in linear regression \cite{tibshirani2005sparsity}, and it encourages $\bbu_i$ and $\bbu_j$ to be equal, thereby forming clusters, while the first term attempts to minimize the within-cluster dissimilarity characterized by the Euclidean distance. In the high-dimensional case, the true underlying clusters may differ only with respect to (w.r.t.) a relatively small number of  features in $\bbx_i$.  While a number of feature selection method have been developed for clustering \cite{alelyani2013feature}, it is desirable to incorporate feature selection and clustering into the same process to possibly improve performance. To this end,  Wang {\it et al.} proposed the sparse convex clustering approach formulated as the following optimization problem \cite{wang2018sparse}: 
\begin{equation}
\label{eq:ccprime}
\begin{split}
    \min_{\bbU \in \mathbb{R}^{p \times n}} \frac{1}{2}\sum_{i=1}^n \lVert\bbx_i - \bbu_i \rVert_2^2 + \lambda \sum_{i < j}\omega_{ij}\lVert\bbu_i - \bbu_j\rVert_q + \gamma \sum_{k=1}^p \nu_{k} \lVert \bba_k \rVert_2,
\end{split}
\end{equation}
where $\gamma>0$ is another tuning parameter, $\nu_{k}$'s are constants that will be specified later, and  $\bba_k^T = [u_{1k}, u_{2k},..., u_{nk}]$ is the $k$th row of $\bbU$, containing the values of the $k$th feature in the $n$ data points. The third term is similar to the group lasso penalty in linear regression \cite{yuan2006model}, and it can shrink certain $\bba_k$'s towards zero, thereby excluding those features from  clustering.  ADMMA and AMA algorithms were developed to solve problems \eqref{eq:cc} and \eqref{eq:ccprime} \cite{chi2015splitting,wang2018sparse}, but they require large computation and memory, and are not applicable to high-dimensional problems with  thousands of data points.

\subsection{Smoothing Proximal Gradient Algorithm}
In this section, we will develop our efficient Sproga algorithm to solve optimization problems \eqref{eq:cc} and \eqref{eq:ccprime}.  Since \eqref{eq:cc} is special case of  \eqref{eq:ccprime} with $\gamma=0$, we will focus on \eqref{eq:ccprime}. For clarity, we will consider the case where $q=2$ in this section, and generalize our algorithm to the case  where $q=1$ or $\infty$ in  \appnorm. 
 
The difficulty of solving \eqref{eq:ccprime} is mainly due to the nonsmooth term $\sum_{i < j}\omega_{ij}\lVert\bbu_i - \bbu_j\rVert_2$. Building on the idea of smooth minimization of non-smooth fuctions \cite{nesterov2005smooth}, we will first find a smooth approximation of this nonsmooth term, and then employ the proximal gradient method to develop our Sproga algorithm. Let us define a set 
$\ccalE=\{(i,j)|\omega_{i,j}\ne 0, i<j\}$. Suppose that the $l$th entry of $\ccalE$ is $(i,j)$, then we  define a matrix $\bbC \in \mathbb{R}^{n \times |\ccalE|}$ as follows:
\begin{equation}
    \bbC_{k, l} = \begin{cases}
    1 \,,\, & k=i\\
    -1 \,,\, &k=j \\ 
    0 \,,\, & {\rm otherwise}
    \end{cases}.
\end{equation}
The second term in  \eqref{eq:ccprime} can be written as $\lambda \sum_{l\in\ccalE}\omega_l\lVert\bbU\bbC_{.,l}\rVert_2$, where we have used $\omega_l$ to represent the $\omega_{i,j}$ associated with the $l$th entry of $\ccalE$, and  $\bbC_{.,l}$ denotes the $l$th column of  $\bbC$. For a given  vector $\bbx$, its Euclidean norm can be written as $\lVert \bbx \rVert_2 = \max_{\lVert\bby\rVert_2 \leq 1} \bby^T \bbx$.  Therefore, we have $\lVert\bbU\bbC_{.,l}\rVert_2= \max_{\lVert\alpha_{l}\rVert_2 \leq 1} \alpha_{l}^T\bbU \bbC_{.,l}$, and we can write \eqref{eq:ccprime} as follows:
\begin{equation}
\label{eq:ccsparse}
\begin{split}
    \min_{\bbU \in \mathbb{R}^{p \times n}} f(\bbU) \equiv \frac{1}{2}\sum_{i=1}^n \lVert\bbx_i - \bbu_i \rVert_2^2 + \lambda \sum_{l \in \ccalE} \omega_{l}  \max_{\lVert\alpha_{l}\rVert_2 \leq 1} \alpha_{l}^T\bbU \mathbb{C}_{.,l} + \gamma \sum_{k=1}^p \nu_{k} \lVert \bba_k \rVert_1.
\end{split}
\end{equation}
Since the second term  in \eqref{eq:ccsparse} is still not a smooth function of $\bbU$,  we will approximate $\max_{\lVert\alpha_{l}\rVert_2 \leq 1} \alpha_{l}^T\bbU \bbC_{.,l}$  with the following function: 
\begin{equation}
\label{eq:gu}
g_{l}(\bbU)=\max_{\lVert\alpha_{l}\rVert_2 \leq 1}  \Bigl(\alpha_{l}^T\bbU \bbC_{.,l} - \frac{\mu}{2} \lVert\alpha_{l}\rVert_2^2 \Bigr),
\end{equation}
where $\mu$ is a small positive constant that controls the approximation error. Since   $\frac{\mu}{2} \lVert\alpha_{l}\rVert_2^2$ is strong convex, and the set $\{\alpha_l:\;\lVert\alpha_{l}\rVert_2 \leq 1\}$ is convex, $g_l(\bbU)$ is convex and continuously differentiable \cite{nesterov2005smooth}.  With this approximation, we can get an approximate solution of \eqref{eq:ccsparse} by solving the following optimization problem: 
\begin{equation}
\label{eq:ccprox}
\begin{split}
    \min_{\bbU \in \mathbb{R}^{p \times n}} \tdf_{\mu}(\bbU) \equiv \frac{1}{2}\sum_{i=1}^n \lVert\bbx_i - \bbu_i \rVert_2^2 + \lambda \sum_{l \in \ccalE} \omega_{l}g_l(\bbU)  + \gamma \sum_{k=1}^p \nu_{k} \lVert \bba_k \rVert_2.
\end{split}
\end{equation} 
Of note, the gap between the approximate function $\tdf_{\mu}(\bbU)$ and the original $f(\bbU)$ is bounded as follows
\begin{equation}
    f(\bbU) - \tdf_{\mu}(\bbU) \leq  \lambda \sum_{l \in \ccalE} \omega_{l} \max_{\lVert\alpha_{l}\rVert_2 \leq 1} \frac{\mu}{2} \lVert\alpha_{l}\rVert_2^2 \leq \frac{\lambda\mu}{2}\sum_{l \in \ccalE} \omega_l.
\end{equation}
If we want to control the approximation within an error of $\epsilon$, we can choose  $\mu = \frac{2\epsilon}{\lambda\sum_{l \in \ccalE} \omega_l}$.

We will next derive a proximal gradient algorithm to solve  \eqref{eq:ccprox}. Let us write $\tdf_{\mu}(\bbU)$ as $\tdf_{\mu}(\bbU)=h(\bbU)+\gamma \sum_{k=1}^p \nu_{k} \lVert \bba_k \rVert_2$, where $h(\bbU) = \frac{1}{2}\sum_{i=1}^n \lVert\bbx_i - \bbu_i \rVert_2^2 + \lambda \sum_{l \in \ccalE} \omega_{l} g_{l}(\bbU)$ is a smooth function, while the second term in $\tdf_{\mu}(\bbU)$ is non-smooth. We need to find the gradient of $h(\bbU)$, but first we need to find the optimal value of $\alpha_l$ in \eqref{eq:gu}, which can be written as $\alpha_l^* = \argmin\limits_{\lVert\alpha_l\rVert_2 \leq 1} \lVert \alpha_l- \frac{\bbU \bbC_{.,l}}{\mu} \rVert_2^2$. It is not difficult to find $\alpha_l^* = \mathbb{P}_2(\frac{\bbU \bbC_{.,l}}{\mu})$, where $\mathbb{P}_2(.)$ is defined as follows:
\begin{equation}
\label{eq:S*operator}
    \mathbb{P}_2(\alpha) = \begin{cases}
    \alpha / \lVert \alpha \rVert_2, & \lVert \alpha \rVert_2 > 1 \\
    \alpha, & \text{otherwise}
    \end{cases}.
\end{equation}

Let us define a matrix $\bbA = [\omega_1\alpha^*_1, \omega_2\alpha^*_2, ..., \omega_{|\ccalE|}\alpha^*_{|\ccalE|}] \in \mathbb{R}^{p \times |\ccalE|}$. Using the results in \cite{nesterov2005smooth} and \cite[Proposition B.25]{bertsekas1999nonlinear}, we can find that the gradient of $\sum_{l \in \ccalE} \omega_{l} g_{l}(\bbU)$ is $\bbA\bbC^T$. Therefore, the gradient of $h(\bbU)$ is given by
\begin{equation}\label{eq:hgrad}
    \nabla h(\bbU) = \bbU - \bbX + \lambda \bbA \bbC^T,
\end{equation}
where $\bbX=[\bbx_1, \cdots, \bbx_n]$. 
Moreover,  $\nabla h(\bbU)$ is Lipschitz continuous with  its Lipschitz constant equal to  $L = 1 + \frac{2\lambda \sum\limits_{l=1}^{|\ccalE|} \omega_l}{\mu}$, as will be proved in \applip.

Let us define the proximal operator for a  lower-semicontinuous convex function $\Omega(\bbz)$  as follows \cite{parikh2014proximal}
\begin{equation}
    \textbf{prox}_{\sigma\Omega}(\bbu) = \argmin_{\bbz} \frac{1}{2}\lVert \bbz - \bbu \rVert_2^2 + \sigma\Omega(\bbz),
\end{equation}
where constant $\sigma>0$.  Then, the proximal gradient method calculates $\bba_k$, $k=1,\cdots, p$, in the $(t+1)$th iteration as follows \cite{parikh2014proximal}:
\begin{equation}\label{eq:aupdate}
    \bba^{t+1}_k = \textbf{prox}_{\frac{\gamma \nu_k}{L}\Omega}\big( \bba_k^t - \frac{1}{L}\nabla h(\bbU^t)_{k,\cdot}\big), k = 1,..., p,
\end{equation}
where $\Omega(\bba_k)=\lVert \bba_k \rVert_2$, and $\nabla h(\bbU^t)_{k,\cdot}$ is the $k$th row of  $\nabla h(\bbU)$ in \eqref{eq:hgrad} evaluated at $\bbU=\bbU^t$.  The proximal operator in \eqref{eq:aupdate} can be explicitly calculated as follows  \cite{parikh2014proximal}:  
\begin{equation}\label{eq:aupdate2}
    \textbf{prox}_{\frac{\gamma \nu_k}{L} \Omega}(\bbu) = \biggl[\Bigl(1 - \frac{\gamma \nu_1}{L\lVert \bbu \rVert_2}\Bigr)_+u_1, \cdots,   \Bigl(1 - \frac{\gamma \nu_n}{L\lVert \bbu \rVert_2}\Bigr)_+u_n  \biggr]^T,
\end{equation}
where  $(x)_+ = \max(0, x)$ and $u_k$ is the $k$th element of $\bbu$.  

Of note, when $\gamma=0$, the proximal gradient method reduces to the gradient descent method which updates $\bba_k$ as follows:  $\bba^{t+1}_k=\bba_k^t - \frac{1}{L}\nabla h(\bbU^t)_{k,\cdot}$. Therefore, our Sproga algorithm is applicable to both \eqref{eq:cc} and \eqref{eq:ccprime}. While we can simply set $v_k=1$, we can also choose $v_k$ as  $1/\lVert \hhatbba_k^{(\gamma = 0)} \rVert_2$,  as suggested in \citep{wang2018sparse}, where $\hhatbba_k^{(\gamma = 0)}$ is the solution  of \eqref{eq:cc}, which can be obtained with the gradient descent method. This choice may improve feature selection. 
The smooth proximal gradient algorithm is summarized in Algorithm 1. Since  $h(\bbU)$ is convex and smooth, we apply the FISTA technique \citep{beck2009fast}, as shown on lines 13 and 14 in Algorithm 1,  to accelerate the convergence of the algorithm.

\begin{algorithm}[htp] \label{alg.sproga}
\caption{Smoothing proximal gradient algorithm (Sproga)}
\begin{algorithmic} [1]
\REQUIRE $\bbX \in \mathbb{R}^{p \times n},\, \lambda, \, \gamma, \, \varepsilon,\, \eta$, maxit,
          $\{\omega_l\}_{l=1}^{|\ccalE|}$, $\{\hhatbba_k^{(\gamma = 0)}\}_{k=1}^p$ \newline
\STATE $\mu = \frac{2\varepsilon}{\lambda\sum_{l=1}^{|\ccalE|}\omega_l}$,
    $\bbnu_{k} = 1 \big/ \lVert \hhatbba_k^{(\gamma = 0)} \rVert_2$,
     $L = 1 + {2\lambda \sum\limits_{l=1}^{|\ccalE|} \omega_l}/{\mu}$ \newline
\STATE  $\bbU^{1} = \bbV^{1} = \bb0$, $t = 1$, $\tau_1 = 1$ \newline
\WHILE{$t < \text{maxit}$ } 
\STATE $\bbA = [\omega_1\alpha^*_1, \omega_2\alpha^*_2, ..., \omega_{|\ccalE|}\alpha^*_{|\ccalE|}]$ with  
  $\alpha^*_l = \mathbb{P}_2(\frac{\bbV^{t} \bbC_{.,l}}{\mu})$ \eqref{eq:S*operator} \newline
\STATE $\nabla h(\bbV^{t}) = \bbV^{t} - \bbX + \lambda \bbA \bbC^T$ \eqref{eq:hgrad}\newline
\STATE $\bbU^{t+1} = \bbV^{t} - \frac{1}{L}\nabla h(\bbV^{t})$ \newline
\FOR{$k=1$ \TO $p$} 
  \STATE       {$\bba^{t+1}_k \leftarrow \textbf{prox}_{\frac{\gamma \nu_k}{L}\Omega}\bigl( \bba_k^{t+1}\bigr)$} \eqref{eq:aupdate}, \eqref{eq:aupdate2}  \newline
\ENDFOR         
\IF{${\lVert\bbU^{t+1} - \bbU^{t}\rVert}/{(1 + \lVert \bbU^{t} \rVert)} \leq \eta$} 
\STATE \textbf{break} \newline
\ENDIF \newline
\STATE $\tau_{t+1} = \frac{1 + \sqrt{1 + 4\tau_t^2}}{2}$\newline
\STATE $\bbV^{t+1} = \bbU^{t+1} - \frac{\tau_t - 1}{\tau_{t+1}}(\bbU^{t+1} - \bbU^{t})$ \newline
\STATE $t = t + 1$ \newline
\ENDWHILE \newline
\RETURN $\bbU^{t}$
\end{algorithmic}
\end{algorithm}

\subsection{Convergence rate and time complexity}
Both optimization problems \eqref{eq:cc} and \eqref{eq:ccprime} are convex, and therefore, the ADMM and AMA algorithms \cite{chi2015splitting,wang2018sparse} converge to the global optimal solution, although it is known that the convergence speed of ADMM is generally low \cite{boyd2011distributed}. It turns out that our Sproga algorithm converges to the global optimal solution at a sublinear rate, as stated formally in the following theorem.  

\begin{theorem}
Let $\bbU^*$ be the optimal solution of \eqref{eq:cc} or \eqref{eq:ccprime} and $\bbU^t$ be the approximate solution produced by Algorithm 1 at the $i$th iteration. Then we have $|f(\bbU^t)-f(\bbU^*)|<O(1/t)$. Moreover, if we require $|f(\bbU^t)-f(\bbU^*)|<\epsilon$ and set $\mu = \frac{\epsilon}{2\lambda\sum_{l \in |\ccalE|} \omega_l}$, then the number of iterations $t$ is upper-bounded by
\begin{equation}\label{eq:tbound}
    \sqrt{ \frac{4\lVert\bbU^*-\bbU^0)\rVert_F^2}{\epsilon}\Bigg(1 + \frac{4\lambda^2\Bigl(\sum\limits_{l=1}^{|\ccalE|} \omega_l\Bigr)^2}{\epsilon}\Bigg) }
\end{equation}
\end{theorem}
The theorem is proved in \apptheorem.

In each iteration of our Sproga algorithm, computing $\nabla h(\bbU)$  requires $O(p|\ccalE|)$ flops, and  updating $\bbU$ with the proximal operator requires $O(np)$ flops. Suppose that we use a k-NN graph to set the weights $\omega_{ij}$, i.e., for  $\bbx_i$, we set $\omega_{ij}\ne 0$ for $j\in \ccalN_k(i)$, where $\ccalN_k(i)$ represents the set of $k$ data points that are nearest to $\bbx_i$.  Then, we have $|\ccalE|=kn$, and the computational complexity per iteration of our Sprogra is $O(knp)$, which is the same as that of AMA \cite{chi2015splitting}. However, our Sproga converges much faster than AMA, and it can be faster than AMA \cite{chi2015splitting} and S-AMA \cite{wang2018sparse} by one to  two orders of magnitude,  as will be shown in simulation results.  

The memory space required by our Sproga is mainly composed of  two parts:  one for $\bbU$, whose size is $O(np)$, and another one for the sparse matrix $\bbC$, whose size is $O(nk)$. Therefore, the total storage required by our Sproga is  $O(n(p+k))$. In AMA and S-AMA, due to the Lagrange multipliers, the total storage  is $O(pn+2pkn)$ \cite{chi2015splitting}. Our simulations show that we typically require $k\ge 10$ for AMA, S-AMA, or Sproga to get good clustering results. Therefore, the storage of AMA and S-AMA is larger than that of our Sproga by at least one order of magnitude.

\subsection{Parameter selection}\label{sec.para}
The weights $\omega_{i,j}, (i,j)\in\ccalE$ can significantly affect the accuracy and the speed of clustering analysis. It was suggested \citep{chi2015splitting}  that $\omega_{ij} =  \exp{(-{\phi}{\lVert \bbx_i - \bbx_j\rVert_2^2})}$ for $j\in \ccalN_k(\bbx_i)$,  where $\phi\ge 0$ and $\ccalN_k(\bbx_i)$ is the set of   $k$ data points that are nearest to $\bbx_i$, and $\omega_{ij}=0$ for $j\notin \ccalN_k(\bbx_i)$. Basically, we can first build a k-NN graph using all data points as nodes of the graph, and then calculate the weight for each edge as $\omega_{ij} =  \exp{(-{\phi}{\lVert \bbx_i - \bbx_j\rVert_2^2})}$. It was suggested that $\phi=0.5$ \cite{wang2018sparse}. This method of weight selection with a $\phi>0$ works well for the datasets where all clusters have a similar density of data points. However, when  different clusters have significantly different density, it may negatively impact the clustering result, because the weights for the data points in low-density clusters may become very small. Based on our empirical results, we  found the following weight selection method is more robust. We first build a k-NN graph. Let us denote the set of edges of the graph as $\ccalE$. For each edge $(i,j)\in \ccalE$, there is a distance $d_{ij}=\lVert \bbx_i - \bbx_j\rVert_2^2$. We rank $d_{ij}$ in the descent order, and remove those edges whose associated distances are in the top 10 percentile. Then, the weights $\omega_{ij}=1$ for those edges remained in $\ccalE$. We name this as the filtered k-NN method. The reason for removing the edges in the top 10 percentile is that those edges most likely connect two data points in two different clusters. Our numerical experiments also showed that when $k\ge 10$, the clustering result is quite robust w.r.t the values of $k$. 

The parameter $\lambda$ determines the number of clusters that the algorithm will output, while the parameter $\gamma$ will determines the number of features selected and may also affect the clustering result. For a given dataset, we need to run the algorithm with a set of values for $\lambda$ and $\gamma$, and then select the best result using. e.g., the gap statistics \cite{tibshirani2001estimating}, although the method of selecting the best clustering result is out of the scope of this paper. In order to select appropriate values of $\lambda$ and $\gamma$, we need to know the range of $\lambda$ and $\gamma$. In \applambda, we derive the maximum and the minimum values of $\lambda$, $\lambda_{\max}$ and $\lambda_{\min}$. In \appgamma, we derive the maximum value of $\gamma$, $\gamma_{\max}$. We then can choose a set of value $\{\lambda_{\max}, \rho_1\lambda_{\max}, \cdots, \rho_1^k\lambda_{\max}\}$ for $\lambda$, where $ 0<\rho_1<1$ and $\rho_1^k\lambda_{\max}\ge \lambda_{\min}$, and a set of values $\{\gamma_{\max}, \rho_2\gamma_{\max}, \cdots, \rho_2^k\gamma_{\max}\}$ for $\gamma$, where $ 0<\rho_2<1$.

\section{Computer Simulation}
In this section, we conduct computer simulation studies to compare the performance of our Sproga algorithm with that of seven other clustering methods, including the convex clustering methods based on the AMA algorithm \citep{chi2015splitting} and the $\text{S-AMA}$  algorithm \citep{wang2018sparse}, K-means, hierarchical clustering, a spectral clustering method named SPECC \citep{ng2002spectral}, a density-based clustering method named DBSCAN \cite{ester1996density}, and a graph-based method Louvain \citep{blondel2008fast}. AMA and S-AMA were implemented with R packages {\it cvxclustr} \citep{chi2015splitting} and {\it scvxclustr} \citep{wang2018sparse}, respectively. K-means and hierarchical clustering were implemented with R package {\it mclust} and a built-in R function {\it hclust}, respectively. SPECC, DBSCA, and Louvain were implemented with R packages {\it kernlab}, {\it dbscan}, and {\it igraph}, respectively. We did not use the convex clustering method based on ADMM, because it is much slower than AMA-based convex clustering, and offers similar performance to that of AMA \cite{chi2015splitting,wang2018sparse}. 

To assess the performance of differnet clustering methods, we computed the adjusted RAND Index (ARI) \cite{hubert1985comparing} and the normalized mutual information (NMI) \cite{studholme1999overlap} by comparing the result of a clustering method with the ground truth cluster assignment. For K-Means,  SPECC, and Louvain, we input the true number of clusters to the algorithms. For hierarchical clustering, we chose the cutoff   values for the distance such that the  number of clusters that the algorithm output was equal to the true number of clusters. We  ran the hierarchical clustering algorithm with three linkage criteria (average, single, and complete linkage), and  chose the best result.  Therefore, the results we will show for K-means, SPECC, Louvain, and hierarchical clustering are the best results of these methods.  For DBSCAN, the optimal value of the parameter epsilon was determined using the method in \citep{rahmah2016determination}. For fair comparison, we ran Sproga  over a set of values of $\gamma$ and $\lambda$ as determined with the steps described in section \ref{sec.para} and used the best result in performance comparison. AMA and S-AMA  were run with the optimal values of $\lambda$ and $\gamma$ determined by Sproga.  We used the filtered k-NN method to determine the weights $\omega_{ij}$ for Sproga, AMA, and S-AMA, and chose $k=50$ for simulation settings 1, 2, and 4, and $k=75$ for simulation setting 3.  Louvain does not need the original data but a graph of the data as input, and we input the same k-NN graph used by our Sproga to Louvain.

In the first simulation setting, we generated  $n = 1,200$ data points evenly distributed in $K = 6$ clusters. The number of samples for the $i$th cluster is apparently $n_i=200$, $\forall i$. The number of features was chosen to be $p = 200$, and only $p_{in}=20$ features were informative for clustering. We used an approach similar to {\it clusterlab} \cite{john2020m3c} to generate these data points. Specifically, we first placed $K$ points evenly on a circle of a radius $r=4$ as the centers of the $K$ clusters.   For the $i$th cluster,  $n_i$ $1\times 2$ vectors were generated from the normal distribution $\ccalN(0, \sigma_i^2\bbI)$, and added to the center of the cluster.  Here, we chose $\sigma_i^2=0.5$, $i=1,\cdots, K$. A matrix $\tdbbX_i$ of $n_i \times 2$ was formed  with these $n_i$ vectors being the rows, and an $n\times 2$ matrix $\tdbbX$ was defined as $\tdbbX=[\tdbbX_1^T,\cdots, \tdbbX_K^T]^T$. We then randomly generated a $2 \times p_{in}$ matrix $\bbW$ whose rows were orthonormal, and computed $\bbarbbX=\tdbbX\bbW$, which represents the data of $p_{in}$ informative features.  For the non-informative features, an  $n\times (p-p_{in})$ matrix $\bbarbbX'$ was obtained with its elements independently generated from the normal distribution $\ccalN(0,\sigma_i^2/2)$. The final simulated data  is  $\bbX=[\bbarbbX, \bbarbbX']$.

Table \ref{tab:set12} reports the performance of eight clustering methods on the simulated data. To ensure a robust comparison, we generated $20$ datasets independently, and ran the clustering algorithms with these datasets. The average ARI and NMI over $20$ datasets and the standard deviation (STD) of ARIs and NMIs are listed in  Table \ref{tab:set12}. All algorithms performed well in this setting, with their average ARI $>0.88$ and average NMI $>0.94$. AMA achieved the perfect result with ARI=1. Louvain and our Sproga offered the second best performance with their average ARI equal to 0.99, while K-means offered the worst performance. The CPU time of eight algorithms for clustering this dataset is also included in Table \ref{tab:set12}.  All algorithms were run on a desktop computer with an i7-5820K (3.30 GHz) CPU and $48$GB RAM with their optimal or default hyper-parameter values.  It is seen that our Sproga is faster than two other convex clustering algorithms, AMA and S-AMA, by two orders of magnitude in this setting. Sproga, AMA, and S-AMA used the same convergence criterion.  Sproga converged at about 300 iterations, AMA converged at 9199 iterations on average, whereas S-AMA reached the default maximum number of iterations (10,000) and was terminated  without satisfying the convergence criterion. K-means and Louvain are much faster than other algorithms on this dataset. The CPU time of SPECC is similar to that of our Sproga in this simulation setting. However, the time complexity of SPECC is $O(n^3)$, which is much larger than that of our Sproga, when $n$ is large.

\begin{table}
\centering
\caption{{\bf Performance of clustering algorithms in simulation settings 1 and 2. }}
\scalebox{0.93}{
\begin{tabular}{|c|c|c|c|c|c|c|}\hline 	
\multirow{2}{*}{Algorithm} & \multicolumn{2}{c|}{Simulation Setting 1} & CPU time & %
    \multicolumn{2}{c|}{Simulation Setting 2} & CPU time \\\cline{2-3}\cline{5-6}
 	& ARI (STD) & NMI (STD) & (secs)~~~~ & ARI  & NMI & (secs)~~~~\\\hline

Sproga	&	0.999(0.001)	&	0.998(0.001) &	9.654	&	{\bf 0.933(0.024)}	&	{\bf 0.918(0.003)}	& 64.803 \\\hline
S-AMA	&	0.973(0.012)	&	0.968(0.013)	&	754.542 &	0.875(0.072)	&	0.865(0.039)	& 1538.530	\\\hline
AMA	&	{\bf 1.000(0.000)}	&	{\bf 1.000(0.000)}	&	985.609  &	0.910(0.055)	&	0.897(0.030)	& 2504.035	\\\hline
K-means	&	0.882(0.124)	&	0.942(0.061)	&	0.042 &	0.699(0.093)	&	0.791(0.047)	& 0.046	\\\hline
SPECC	&	0.888(0.117)	&	0.945(0.051)	&	7.570 &	0.799(0.154)	&	0.864(0.080)	& 5.482	\\\hline
DBSCAN	&	0.979(0.056)	&	0.989(0.023)	&	0.273 &	0.777(0.195)	&	0.794(0.108)	& 0.376	\\\hline
Louvain	&	0.999(0.001)	&	0.999(0.001)	&	0.045 &	0.799(0.116)	&	0.867(0.034)	& 0.140	\\\hline
Hierarchical	&	0.929(0.092)	&	0.972(0.036)	&	1.088 &	0.880(0.092)	&	0.871(0.082)	& 0.918	\\\hline

\end{tabular}
}
\label{tab:set12}
\end{table}

In the second simulation setting,  we generated  $n = 1,000$ data points unevenly distributed in $K = 6$ clusters. The number of samples in six clusters are $20, 60, 120, 100, 300,$ and  $400$. The number of features was again chosen to be $p = 200$, among which $p_{in}=20$ features were informative. We used the same procedure described in setting 1 to generate $n=1,000$ data points. The difference was that the variances $\sigma_i^2$, $i=1,\cdots, K$, were different for different clusters, and they were generated from positive values of the normal distribution $\ccalN(1, 1)$.  Again, we generated 20 independent datasets and ran eight algorithms with these datasets. The results for this simulation setting is also included in Table \ref{tab:set12}. Our Sproga significantly outperforms other algorithms. The CPU time of AMA and S-AMA is more than 20 times of the CPU time of Sproga. Sproga converged at 2,400 iterations on average, whereas AMA and S-AMA did not satisfy the convergence criterion after 25,000 iterations and were terminated. The observations in both settings 1 and 2 show that convergence speed of AMA and S-AMA is much lower than that of Sproga.

\begin{figure}[ht]
\centering
\includegraphics[width=0.6\textwidth]{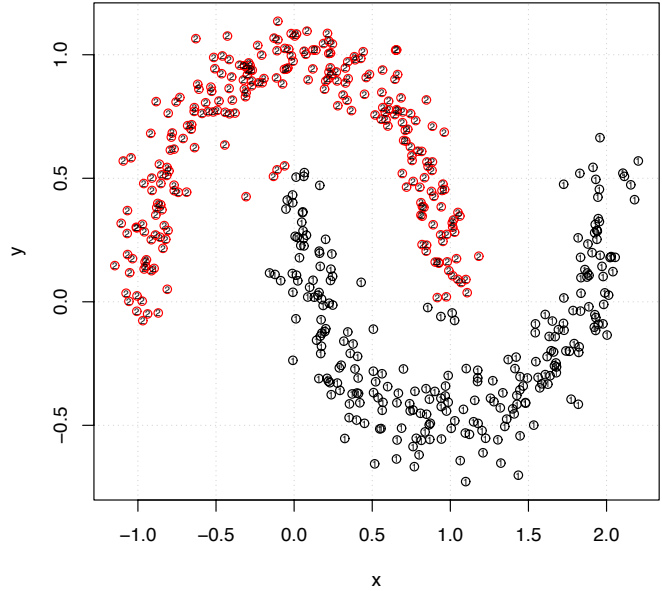}
\caption{Visualization of the $1,000$ data points in simulation setting 3 with their first two features. }
\label{fig1-moon}
\end{figure}

In the third simulation setting, we generated a non-Gaussian dataset in the form of two interlocking half-moons. Each simulated dataset contains $n = 1,000$ observations in $K = 2$ clusters with $p = 200$ features. We first adopted the function {\it sklearn.datasets.make\_moons} in the Python machine learning library Scikit-Learn to generate a 2D half-moon dataset. More specifically, 2D points were generated as $(|r \cos(\theta_i)|, r\sin(\theta_i)) + \ccalN(0, \sigma^2)$ for the first half-moon, and $(a - |r\cos(\theta_j)|, b - r\sin(\theta_j))+  \ccalN(0, \sigma^2)$ for the second half-moon, where $\theta_i, \theta_j \in (0, \pi)$. Each cluster contained $500$ points, and we set $r = 1$, $a = 1$, $b = 0.5$, and $\sigma^2 = 0.1$. To generate high-dimensional data points, we randomly generated $p-2$ non-informative features from normal distribution $\ccalN(0, 0.01)$ and concatenated them to the 2D data. We generated 20 datasets independently, and ran eight clustering algorithms with these datasets. One of the 20 datasets is shown in Figure \ref{fig1-moon}.  The performance of the clustering algorithms on these datasets is described in Table \ref{tab:set34}.  Sproga, AMA, S-AMA, SPECC, and DBSCAN offered similar performance, and they significantly outperformed K-means, Louvain, and hierarchical clustering. 

\begin{table}
\centering
\caption{{\bf Performance of clustering algorithms in simulation settings 3 and 4.}}
\begin{tabular}{|c|c|c|c|c|r|}\hline 	
\multirow{2}{*}{Algorithm} & \multicolumn{2}{c|}{Simulation Setting 3} & %
    \multicolumn{2}{c|}{Simulation Setting 4} \\\cline{2-5}
 	& ARI (STD) & NMI (STD) & ARI  & NMI  \\\hline 	
 	
Sproga	&	0.944(0.035)	&	0.937(0.079)	&	{\bf 0.758}	&	{\bf 0.726}	\\\hline
S-AMA 	&	0.922(0.010)	&	0.918(0.034)	&	–	&	–	\\\hline
AMA 	&	{\bf 0.955(0.015)}	&	{\bf 0.944(0.025)}	&	–	&	–	\\\hline
K-means	&	0.256(0.020)	&	0.195(0.016)	&	0.587	&	0.689	\\\hline
SPECC	&	0.945(0.034)	&	0.901(0.051)	&	0.581	&	0.649	\\\hline
DBSCAN	&	0.895(0.019)	&	0.847(0.016)	&	0.397	&	0.510	\\\hline
Louvain	&	0.158(0.011)	&	0.545(0.010)	&	0.440	&	0.675	\\\hline
Hierarchical 	&	0.405(0.086)	&	0.433(0.087)	&	0.719	&	0.692	\\\hline

\end{tabular}
\label{tab:set34}
\end{table}

\begin{table}
\centering
\caption{{\bf Accuracy of feature selection for Sproga and S-AMA algorithms.}}
\begin{tabular}{|c|c|c|c|c|r|}\hline 	
\multirow{2}{*}{Simulation Setting} & \multicolumn{2}{c|}{Sproga} & %
    \multicolumn{2}{c|}{S-AMA} \\\cline{2-5}
 	& PD (STD) & FDR & PD (STD)  & FDR  \\\hline 	
1	&	0.970(0.035)	&	0.000	&	0.980(0.026)	&	0.000	\\\hline
2	&	0.960(0.022)	&	0.000	&	0.960(0.022)	&	0.000	\\\hline
3	&	1.000(0.000)	&	0.000	&	1.000(0.000)	&	0.000	\\\hline
4	&	0.980	&	0.000	&	–	&	–	\\\hline
\end{tabular}
\label{tab:pd}
\end{table}

\begin{figure}[ht]
\centering
\includegraphics[width=0.6\textwidth]{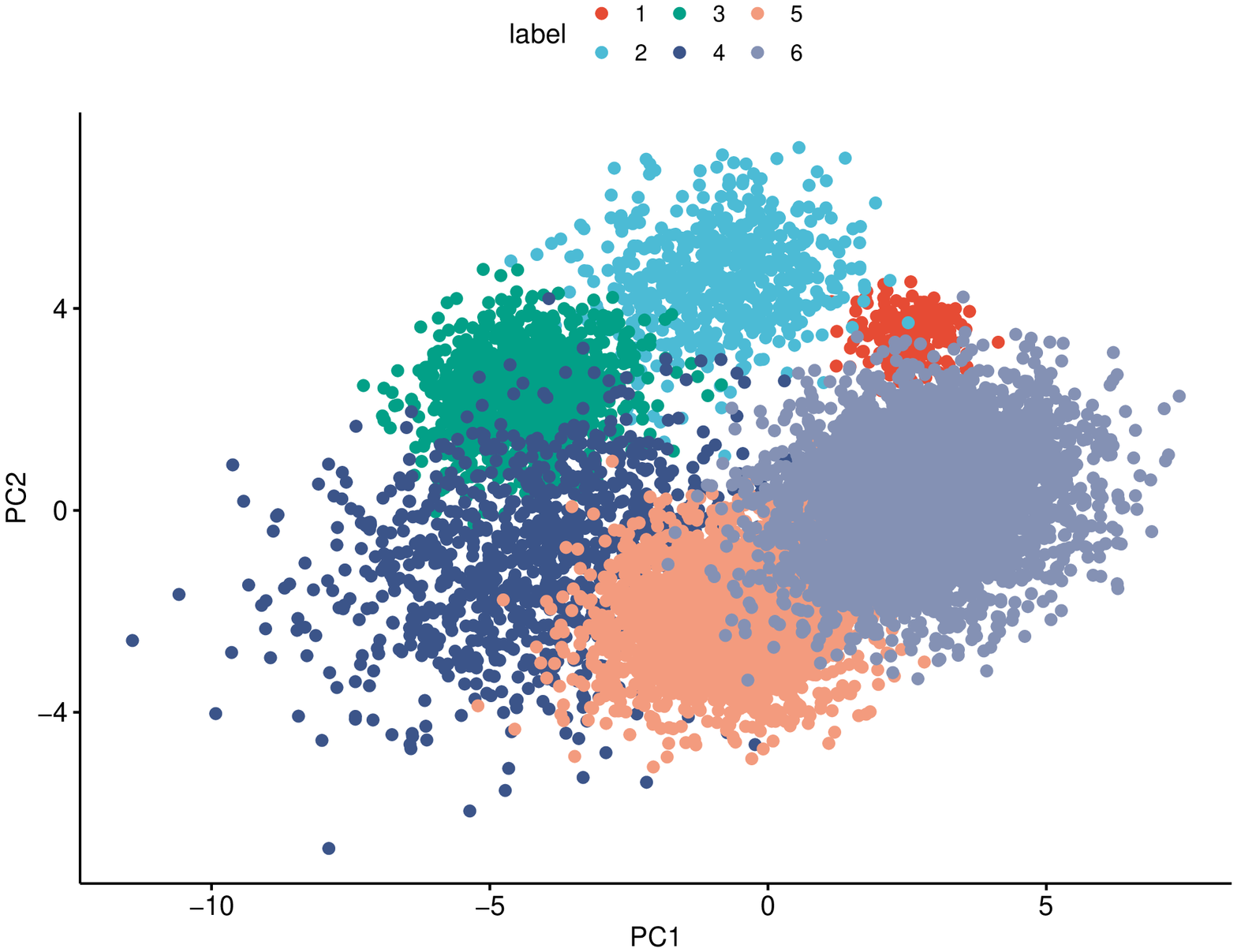}
\caption{Visualization of $10,000$ data points in simulation setting 4 with the first and second principal components of the data matrix.}
\label{fig1b-largeNP}
\end{figure}

In the fourth simulation setting, we tested performance of eight clustering algorithms when the number of data points was large. We used the same procedure as the one in simulation setting 2 to generate data with the following changes: $n=10,000$, $p=500$, $p_{inf}=100$, and the number of data points in $K$ clusters are 200, 600, 1200, 1000, 3000, and 4000, respectively.  Performance of five algorithms is included in Table \ref{tab:set34}. AMA and S-AMA were very slow; they eventually used up the 48GB memory space and 22GB swap space on the disk, and failed. Our Sproga again significantly outperformed K-means, SPESCC, DBSCAN, Louvain, and hierarchical clustering. The $10,000$ data points are visualized in Figure \ref{fig1b-largeNP} with the first two principal components (PCs) of the data matrix. It is seen that six clusters are not separated well on two PCs. This may be the reason why  all  algorithms except Sproga and hierarchical clustering achieved an ARI less than 0.6, although our Sproga achieved the highest ARI of 0.785. 

As described in the method section, similar to S-AMA \cite{wang2018sparse}, our Sproga algorithm can perform feature selection and clustering simultaneously. Table \ref{tab:pd} shows the performance of feature selection for both Sproga and S-AMA. After we used Sproga or S-AMA to complete clustering of a dataset, we identified the features that were not shrunk to zero. We compared these selected features with the true informative features used in the generation of the data, and computed the power of detection (PD) and the false discovery rate (FDR). Recall that in simulation settings 1, 2, and 3, we generated 20 datasets. For each dataset, we obtained a PD and a FDR, and the results in Table \ref{tab:pd} are the average PD and FDR and their standard deviation. It is seen from Table \ref{tab:pd} that PD is equal or very close to one, while FDR is always zero in the four simulation settings. This tells that both Sproga and S-AMA can select features reliably.

\begin{table}
\centering
\caption{{\bf Performance of clustering algorithms on two scRNA-seq data.}}
\begin{tabular}{|c|c|c|c|c|}\hline
\multirow{2}{*}{Algorithm} & \multicolumn{2}{c|}{Mouse Data} & %
    \multicolumn{2}{c|}{Human PBMC Data} \\\cline{2-5}
 	& ARI  & NMI & ARI  & NMI  \\\hline
 	
Sproga	&	{\bf 0.791}	&	{\bf 0.853}	&	{\bf 0.608}	&	{\bf 0.697}	\\\hline
K-means	&	0.419	&	0.596	&	0.479	&	0.578	\\\hline
SPECC	&	0.205	&	0.294	&	0.428	&	0.544	\\\hline
DBSCAN	&	0.552	&	0.733	&	0.047	&	0.113	\\\hline
Louvein	&	0.440	&	0.675	&	0.533	&	0.617	\\\hline
Hierarchical	&	0.382	&	0.446	&	0.341	&	0.499	\\\hline

\end{tabular}
\label{tab:real}
\end{table}

\begin{figure}[ht]
\centering
\includegraphics[width=0.8\textwidth]{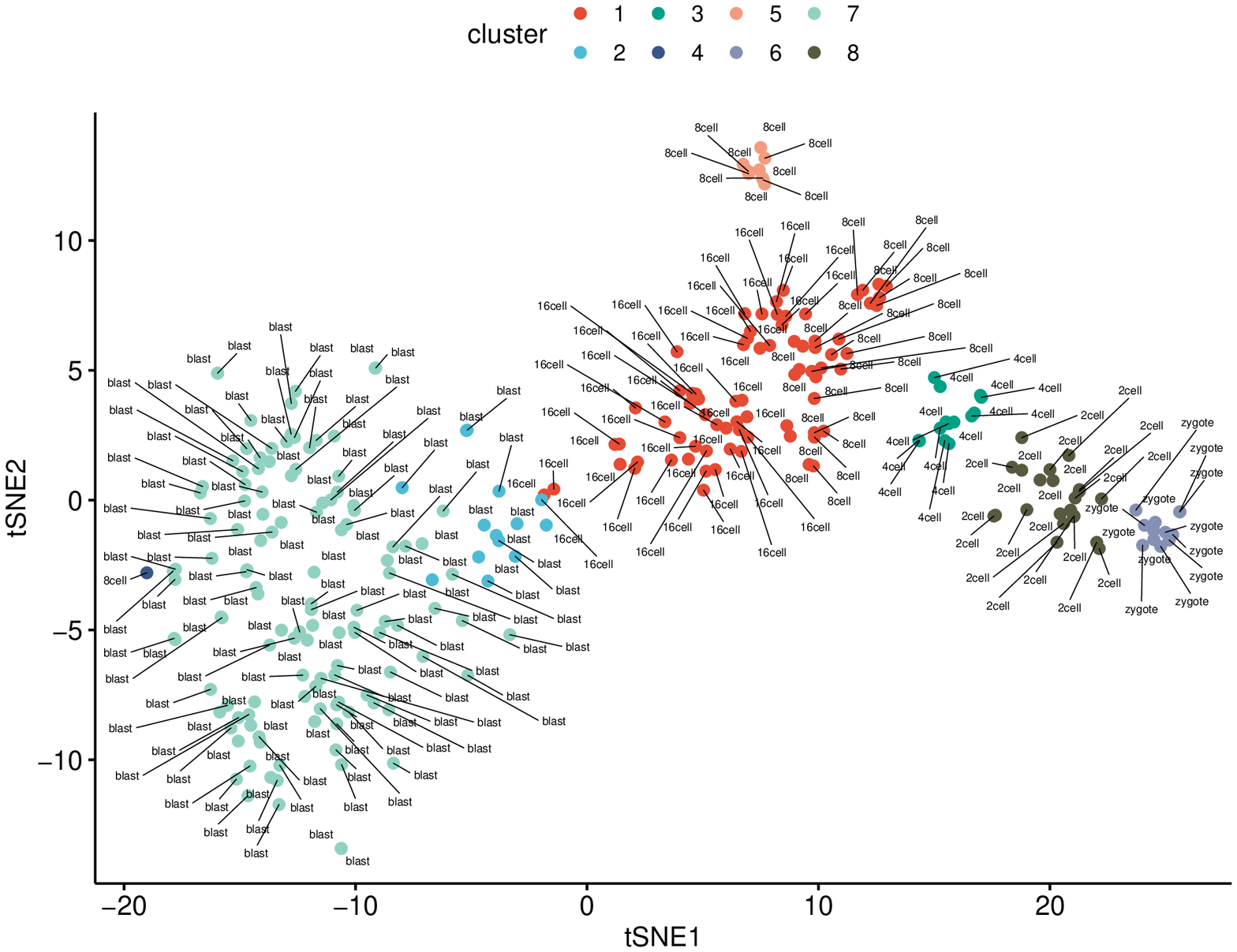}
\caption{Visualization of the clustering result of Sproga with t-SNE for the mouse scRNA-seq data. Each circle represents a cell, labels of circles are the ground truth cell types, and eight colors represent 8 clusters output from Sproga.}
\label{fig:rpkm}
\end{figure}`

\section{Real Data Analysis}
\subsection{Clustering with mouse single cell RNA-seq (scRNA-seq) data}
We compared performance of clustering algorithms using a scRNA-seq dataset of mouse embryonic cells \citep{deng2014single}. The dataset contains the expression values of $22,958$ genes in $268$ individual cells. These cells are classified into $6$ different types resulted from different  development stages: zygote ($12$), 2-cell ($22$), 4-cell ($14$), 8-cell ($37$), 16-cell ($50$) and blast ($133$). The RPKM values of gene expression were pre-processed with the steps described in a recent benchmark study of clustering methods for scRNA-seq data \citep{krzak2019benchmark}. Specifically, all non-expressed genes whose total RPKMs across $268$ cells were $0$ were filtered out. This resulted in $21,174$ genes whose expression levels were used as features in clustering analysis.  The RPKM values of different genes in each cell were normalized by the sum of RPKM values of all genes in the cell, and the normalized RPKM values were transformed to the log-scale. 

Clustering analysis was performed on the processed gene expression data. K-means, SPECC, and Louvain require the number of cluster as their parameter. We input the true $K=6$ to these algorithms. 
For DBSCAN, the optimal value of parameter epsilon was determined  from the elbow of the pairwise distance curve \citep{rahmah2016determination}, and another parameter minPts was set to $5$. For fair comparison, our Sproga searched over a set of values for two parameters $\lambda$ and $\gamma$, and used the best values to obtain the final clustering result. To set the weights $\omega_{ij}$'s, we used a filtered k-NN graph with $k=10$. Louvain used the same k-NN graph to perform clustering. The performance of these clustering methods is shown in Table \ref{tab:real}. AMA and S-AMA failed on this dataset, because the number of features is very large, although the number of data points is relatively small. 
It is seen from Table \ref{tab:real} that our Sproga significantly outperforms other methods. The clustering result of Sproga is visualized in Figure \ref{fig:rpkm} with t-SNE \citep{maaten2008visualizing}. Sproga output 8 clusters, although the true number of clusters is 6.  It is seen from Figure \ref{fig:rpkm} that most errors occur in clusters 1 and 2. In cluster 1, the majority of cells are 16-cell, but a number of cells of 8-cell are included in this cluster, and these cells of 8-cell seem closer to the cells of 16-cell than to other cells of 8-cell in cluster 5. 
Cluster 2 mainly contains cells of 16-cell, which means that cluster 2 should merge with cluster 1. However, it seems that cells in cluster 2 have a large distance to the cells in cluster 1. 

\begin{figure}[ht]
\centering
\includegraphics[width=0.8\textwidth]{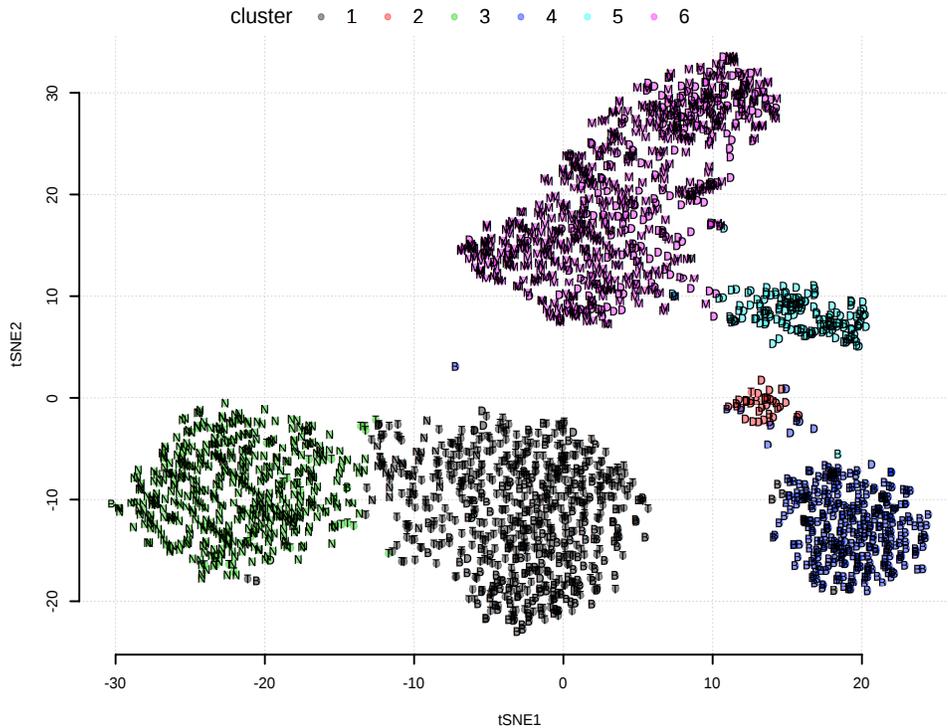}
\caption{Visualization of the clustering result of Sproga with t-SNE for the human PBMC scRNA-seq data. Each circle represents a cell. The label of a circle represents the true cell type  (B: B cell, D: dendritic cell, M: monocyte, N: NK cell, T: T cell).  Six different colors represent the 6  clusters output from Sproga. }
\label{fig:umi}
\end{figure}

\subsection{Clustering with scRNA-seq data of human peripheral blood mononuclear cells (PBMCs)}
We next evaluate the performance of cluster algorithms on the 68k PBMC scRNA-seq dataset \citep{zheng2017massively}. The dataset contains the counts of  unique molecular identifiers (UMIs)  as expression values of $32,738$ transcripts in about 68,000 cells of 11 types including B cell, dendritic cell,  monocyte, NK cell, CD34+ cell, and 6 types of T cell (cytotoxic, helper, memory, naive, naive cytotoxic, and regulatory). Since the number of CD34+ cells is small (277), they were excluded from clustering analysis.  The expression profiles of T cells are similar, and therefore, we treat all T cells as the same type so that the dataset contains 5 cell types. Following the preprocessing procedure in \citep{zheng2017massively}, we removed the transcripts whose summarized counts across all cells are smaller than $1$, and normalized the UMI counts of each gene in a cell by the total UMI counts in the cell. We then calculated the dispersion (variance/mean) of the UMI counts of each gene, and selected the top $1,000$ genes with highest dispersion  as the features for clustering. We randomly selected $500$ cells  from each cell type, which resulted in $2,500$ cells and a $2,500\times 1,000$ data matrix of gene expression values that was analyzed by clustering algorithms. 

Similar to the analysis of the mouse data, we input the true number of clusters $K=5$ to K-means, SPECC, and Louvain, and used the same method to determine parameters of DBSCAN. Our Sproga searched over a set of values for $\lambda$ and $\gamma$ to find the optimal values, and then performed clustering analysis with the optimal parameter values. To set weights $\omega_{ij}$, we used a filtered k-NN graph with $k=50$. Louvain used the same k-NN graph used by Sproga. The performance of five clustering algorithms is shown in Table \ref{tab:real}. Because of the relatively large number of data points and features, AMA and S-AMA failed to obtain any result. As seen from Table \ref{tab:real}, our Sproga again significantly outperforms the other methods.  The clustering result of Sproga is visualized in Figure \ref{fig:umi} with t-SNE. Sproga output 6 clusters, although the true number of clusters is 5. As seen from Figure \ref{fig:umi}, clusters 5 and 6 both contain dendritic cells, but they well separated. Therefore, our Sproga identified them as two separate  clusters.

\section{Conclusion}
In this paper, we have developed a very efficient  smoothing proximal gradient method for sparse convex clustering. Our analysis and simulations showed that our Sproga algorithm could be faster than the state-of-the-art convex clustering algorithms AMA \citep{chi2015splitting} and S-AMA \cite{wang2018sparse} by one to two orders of magnitude, and required much less memory space. Our simulation study and real data analysis also demonstrated the superior performance of our Sproga algorithm over several other well-known clustering methods including K-means, hierarchical clustering, spectral clustering, DBSCAN, and Louvain. The efficiency and superior performance of our algorithm will help convex clustering to find its wide application.

\appendix
\appendixpage
\section{Appendix A: Derivation of the gradient Lipschitz constant}
We will prove that $\nabla h(\bbU)$ in \eqref{eq:hgrad} satisfies $\lVert \nabla h(\bbU) - \nabla h(\bbV) \rVert \le L \lVert  \bbU-\bbV \rVert$ ,  and find an expression for the Lipschtiz constant $L$.  Let us denote $\beta_k^*$ as the optimal solution of the optimization problem $\max_{\lVert\bbbeta_k\rVert_2 \leq 1} \bbbeta_k^T\bbV \bbC_{.,k}- \frac{\mu}{2} \lVert \bbbeta_k\rVert_2^2$, $k = 1, 2, ..., |\ccalE|$, and define a matrix $\bbB = [\bbbeta_1^*, \bbbeta_2^*, ..., \bbbeta_{|\ccalE|}^*]$. Recall that we have defined a similar matrix $\bbA = [\bbalpha_1^*, \bbalpha_2^*, ..., \bbalpha_{|\ccalE|}^*]$ containing the optimal solutions of  $\max_{\lVert\bbalpha_{k}\rVert_2 \leq 1} \bbalpha_{k}^T\bbU \mathbb{C}_{.,k} - \frac{\mu}{2} \lVert\bbalpha_{k}\rVert_2^2 $,  $k = 1, 2, ..., |\ccalE|$.  From \eqref{eq:hgrad}, we have the following:  $\lVert \nabla h(\bbU) - \nabla h(\bbV) \rVert = \lVert \bbU - \bbV + \lambda (\bbA - \bbB) \bbC^T\rVert$, which can also be written as
\begin{equation}\label{eq:lip1}
    \lVert \nabla h(\bbU) - \nabla h(\bbV) \rVert = \lVert \bbU - \bbV + \lambda \sum_{l=1}^{|\ccalE|} \omega_l (\bbalpha_l^* - \bbbeta_l^*)\bbC_{.,l}\rVert. 
\end{equation}
This implies the following
\begin{equation}\label{eq:lip2}
    \lVert \nabla h(\bbU) - \nabla h(\bbV) \rVert \leq \lVert \bbU - \bbV\rVert + \lambda \sum_{l=1}^{|\ccalE|} \omega_l \lVert \bbC_{.,l} \rVert \times \lVert \bbalpha_l^* - \bbbeta_l^* \rVert.
\end{equation}

Let us define two functions $f(\bbalpha_k) =  \bbalpha_{k}^T\bbU \bbC_{.,k} - \frac{\mu}{2} \lVert\bbalpha_{k}\rVert_2^2$ and $g(\bbbeta_k) =  \bbbeta_k^T\bbV \bbC_{.,k}- \frac{\mu}{2}\lVert \bbbeta_k\rVert_2^2$, where constant $\mu > 0$. Since $f(\bbalpha_k)$ and $g(\bbbeta_k)$ are concave, and we have  $\bbalpha_k^*=\arg\max_{\bbalpha_k} f(\bbalpha_k)$ and  $\bbbeta_k^* =\arg\max_{\bbbeta_k} g(\bbbeta_k)$ as defined earlier,  the first order optimality condition implies the following 
\begin{equation}\label{eq:optimal1}
\begin{split}
    \langle \nabla f(\alpha_k^*), \bbbeta_k^* - \bbalpha_k^*\rangle \leq 0,\\
    \langle \nabla g(\beta_k^*),  \bbalpha_k^* - \bbbeta_k^* \rangle \leq 0.
\end{split}
\end{equation}
Since $\nabla f(\alpha_k)=\bbU \bbC_{.,k} - \mu \bbalpha_k$ and $\nabla g(\beta_k)=\bbV \bbC_{.,k} - \mu \bbbeta_k$, \eqref{eq:optimal1} can be written as 
\begin{equation}\label{eq:optimal2}
\begin{split}
    \langle \bbU \bbC_{.,k} - \mu \bbalpha_k^*, \bbbeta_k^* - \bbalpha_k^*\rangle \leq 0,\\
    \langle \bbV \bbC_{.,k} - \mu \bbbeta_k^*, \bbalpha_k^* - \bbbeta_k^* \rangle \leq 0.
\end{split}
\end{equation}
Adding the two equations in \eqref{eq:optimal2}, we obtain 
\begin{equation}
    \langle (\bbU - \bbV)\bbC_{.,k} + \mu(\bbbeta_k^* - \bbalpha_k^*) , \bbbeta_k^* - \bbalpha_k^*\rangle \leq 0,  
\end{equation}
which results in 
\begin{equation}\label{eq:ab}
    \lVert\bbbeta_k^* - \bbalpha_k^* \rVert_2 \leq \frac{\lVert \bbC_{.,k} \rVert_2}{\mu} \lVert \bbV - \bbU \rVert_2.
\end{equation}
%
%
Combining \eqref{eq:lip2}  and and \eqref{eq:ab} and  using $\lVert \bbC_{.,l} \rVert_2 =  \sqrt{2}, \forall l$,  we have
\begin{equation}
    \lVert \nabla h(\bbU) - \nabla h(\bbV) \rVert \leq \biggl(1+\frac{2\lambda\sum\limits_{k=1}^{|\ccalE|} \omega_k}{\mu}\biggr)\lVert \bbU - \bbV\rVert.
\end{equation}
Therefore, $\nabla h(\bbU)$ is Lipschitz continuous with the Lipschitz constant  $L = 1+\frac{2\lambda\sum\limits_{k=1}^{|\ccalE|} \omega_k}{\mu}$.

\section{Appendix B: Proof of Theorem 1} \label{appendix.thoerem1}
Recall that we write the objective function in \eqref{eq:ccprox}  as $\tdf_{\mu}(\bbU)=h(\bbU)+\gamma\sum_{k=1}^p \nu_{k} \lVert \bba_k \rVert_2$. The smooth function $h(\bbU)$ is convex and its gradient is Lipschitz continous with the Lipschitz constant $L$, as we proved in \applip. The second term $\sum_{k=1}^p \nu_{k} \lVert \bba_k \rVert_2$ is convex. Therefore, the solution to \eqref{eq:ccprox} yielded by the Sproga algorithm, which is basically a proximal gradient method,  at the $t$th iteration has the following convergence property \cite{bubeck2017convex}:
\begin{equation}
\begin{split}
    \tdf(\bbU^t) - \tdf(\bbU^*) \leq \frac{2L\lVert\bbU^*-\bbU^0\rVert_F^2}{t^2},
\end{split}
\end{equation}
where $\bbU^*$ is the optimal solution to \eqref{eq:ccprox}. Using this property, we have the following result for the objective function in \eqref{eq:ccsparse} or \eqref{eq:ccprime}:
\begin{equation} \label{eq:deltafU1}
\begin{split}
    f(\bbU^t) - f(\bbU^*) &= (f(\bbU^t) - \tdf(\bbU^t)) + (\tdf(\bbU^t) - \tdf(\bbU^*)) + (\tdf(\bbU^*) - f(\bbU^*)) \\
    &\leq \frac{ \lambda \mu \sum\limits_{l=1}^{|\ccalE|} \omega_l }{2} + \frac{2L\lVert\bbU^*\rVert_F^2}{t^2} + \frac{ \lambda \mu \sum\limits_{l=1}^{|\ccalE|} \omega_l }{2}\\
    &=\lambda \mu \sum\limits_{l=1}^{|\ccalE|} \omega_l + \frac{2\lVert\bbU^*\rVert_F^2}{t^2}\bigg(1 + \frac{2\lambda \sum\limits_{l=1}^{|\ccalE|} \omega_l}{\mu}\bigg),
\end{split}
\end{equation}
where $|\ccalE| = \frac{n(n-1)}{2}$ for a fully connected graph and $|\ccalE| = kn$ for a k-NN graph. From the definition of $\mu$, we have $\epsilon = 2\lambda \mu \sum\limits_{l=1}^{|\ccalE|} \omega_l$, and therefore, \eqref{eq:deltafU1} becomes
\begin{equation}\label{eq:deltafU2}
\begin{split}
    f(\bbU^t) - f(\bbU^*) \leq \frac{\epsilon}{2} + \frac{2\lVert\bbU^*\rVert_F^2}{t^2}\bigg(1 + \frac{4\lambda^2(\sum\limits_{l=1}^{|\ccalE|} \omega_l)^2}{\epsilon}\bigg).\\
\end{split}
\end{equation}
Let the right hand side of \eqref{eq:deltafU2} be equal to $\epsilon$, we obtain the following
\begin{equation} \label{eq:tepsilon1}
    \frac{2\lVert\bbU^*\rVert_F^2}{t^2}\big(1 + \frac{4\lambda^2\Big(\sum\limits_{l=1}^{|\ccalE|} \omega_l\Big)^2}{\epsilon}\big) = \frac{\epsilon}{2},
\end{equation}
which results in the upper bound on $t$ given in \eqref{eq:tbound}. Since $\epsilon \ll 1$, we have 
$1< \frac{4\lambda^2\Big(\sum\limits_{l=1}^{|\ccalE|} \omega_l\Big)^2}{\epsilon}$. Therefore, we obtain the following inequality from \eqref{eq:tepsilon1}:
\begin{equation} \label{eq:tepsilon2}
    \frac{4\lVert\bbU^*\rVert_F^2}{t^2} \times \frac{4\lambda^2\Big(\sum\limits_{l=1}^{|\ccalE|} \omega_l\Big)^2}{\epsilon}> \frac{\epsilon}{2},
\end{equation}
which results in  
\begin{equation}
        \epsilon < \frac{4\sqrt{2}\lambda\lVert\bbU^*\rVert_F\sum\limits_{l=1}^{|\ccalE|} \omega_l}{t}.
\end{equation}
 Therefore, we have $f(\bbU^t) - f(\bbU^*) < \frac{4\sqrt{2}\lambda\lVert\bbU^*\rVert_F\sum\limits_{l=1}^{|\ccalE|} \omega_l}{t}$.
Since $\lambda$, $\lVert\bbU^*\rVert_F$, and $\omega_l$ are fixed constants, we get the $O(1/t)$ convergence rate.

\section{Appendix C: Sproga algorithm for $\ell_1$- or $\ell_{\infty}$-norm}
In \eqref{eq:ccprime}, we considered $\ell_q$-norm $\lVert \bbu_i-\bbu_j \rVert_q$, where $q=1, 2, \cdots, \infty$. However, we assumed $q=2$ in the subsequent derivation of the Sproga algorithm.  Here, we will extend Sproga  to the  case where $q=1$ or $\infty$.  

For a vector $\bbx$, its $\ell_q$-norm can be written as 
\begin{equation}
    \lVert \bbx \rVert_q = \max_{\lVert \bby \rVert_s \leq 1} \bby^T\bbx \,,\;\;\; \frac{1}{s} + \frac{1}{q} = 1.
\end{equation}
Using this formulation of the $\ell_q$-norm, we can write \eqref{eq:ccprime} as 
\begin{equation}
\label{eq:gccdual}
\begin{split}
    \min_{\bbU \in \mathbb{R}^{p \times n}} f(\bbU) \equiv \frac{1}{2}\sum_{i=1}^n \lVert\bbx_i - \bbu_i \rVert_2^2 + \lambda \sum_{l \in \ccalE} \omega_{l}  \max_{\lVert\alpha_{l}\rVert_s \leq 1} \alpha_{l}^T\bbU \bbC_{.,l} + \gamma \sum_{k=1}^p \nu_{k} \lVert \bba_k \rVert_2,
\end{split}
\end{equation}
where $s=1/(1-1/q)$. We can approximate $\max_{\lVert\alpha_{l}\rVert_s \leq 1} \alpha_{l}^T\bbU \bbC_{.,l}$ with the following function
\begin{equation}\label{eq:gu2}
g_l(\bbU)= \max_{\lVert\alpha_{l}\rVert_s \leq 1} \Big( \alpha_{l}^T\bbU \bbC_{.,l} - \frac{\mu}{2} \lVert\alpha_{l}\rVert_2^2\Big),
\end{equation}
where the small positive constant $\mu$ controls the approximation error. Since   $\frac{\mu}{2} \lVert\alpha_{l}\rVert_2^2$ is strong convex, and the set $\{\alpha_l:\;\lVert\alpha_{l}\rVert_s \leq 1\}$ is convex, $g_l(\bbU)$ is convex and continuously differentiable \cite{nesterov2005smooth}. With this new definition of $g_l(\bbU)$ instead of the one in \eqref{eq:gu}, we still can use \eqref{eq:ccprox} to find an approximate solution to \eqref{eq:ccprime}, and we only need to modify the fouth line of the Sproga algorithm by changing $\alpha_l^*$ to be the optimal solution of \eqref{eq:gu2}. 

We next find $\alpha_l^*$ from \eqref{eq:gu2} for different values of $s$ or equivalently $q$. The optimal solution of \eqref{eq:gu2} can be written as
\begin{equation}
    \alpha_l^* = \argmin\limits_{\lVert\alpha_l\rVert_s \leq 1} \biggl\lVert \alpha_l- \frac{\bbU \bbC_{.,l}}{\mu} \biggr\rVert_2^2 =  \mathbb{P}_s\biggl(\frac{\bbU \bbC_{.,l}}{\mu}\biggr), 
\end{equation} 
where $\mathbb{P}_s(\cdot)$ has been defined in \eqref{eq:S*operator} for $s=2$, and we need to find its explicit expression for other values of $s$.  We denote the $i$th element of $\mathbb{P}_s(\cdot)$ as $[\mathbb{P}_s(\cdot)]_i$.

Let us first consider the case where $q=1$ and $s=\infty$.  In this case, it is not difficult to find the following
\begin{equation}
    [\mathbb{P}_{\infty}(\bbz)]_i = \begin{cases}
    -1,  & z_i < -1 \\
    ~~~1 ,   & z_i > 1 \\
    ~~~z_i,  & |z_i| \leq 1
    \end{cases},
\end{equation}
where $z_i$ is the $i$th element of vector $\bbz$. 

Let us next consider the case where $q=\infty$ and $s=1$. In this case, we have $\mathbb{P}_{1}(\bbz)= \argmin_{\lVert\bbx\rVert_1 \leq 1} \frac{1}{2} \lVert\bbx - \bbz\rVert_2^2$. The Lagrangian of this optimization problem is  
\begin{equation}
    L(\bbx, \lambda) = \frac{1}{2}\lVert \bbx - \bbz \rVert_2^2 + \lambda (\lVert \bbx \rVert_1 -1).
\end{equation}
From the stationarity condition of KKT conditions, we  have $\nabla_{\bbx}L(\bbx, \lambda) = 0$, which implies the following
\begin{equation}\label{eq:kktstat}
   x_i-z_i+\lambda s_i=0, i=1,\cdots, p,
\end{equation}
where $s_i$ is the subgradient of $|x_i|$; $s_i=1$ if $x_i>0$, $s_i=-1$ if $x_i<0$, and $x_i\in [-1, 1]$ if $x_i=0$. From the complementary slackness  condition     $\lambda (\lVert \bbx \rVert_1 - 1) = 0$, we know that $\lambda=0$ if $\lVert \bbx \rVert_1 <1$, which together with \eqref{eq:kktstat}
implies $\bbx=\bbz$ if $\lVert \bbz \rVert_1 <1$. 

If $\lVert \bbx \rVert_1 =1$, then $\lambda>0$, and \eqref{eq:kktstat} yields the following
\begin{equation}
\begin{split}
    x_i = \mathbb{S}(z_i,\lambda )=\begin{cases}
    z_i + \lambda, & z_i < -\lambda \\
    0,             & |z_i| \leq \lambda \\
    z_i - \lambda, & z_i > \lambda
    \end{cases}.
\end{split}
\end{equation}
Note that $\mathbb{S}(z_i,\lambda )$ can also be written as $\mathbb{S}(z_i,\lambda )=
\text{sgn}(z_i)(|z_i| - \lambda)_+$, where $\text{sgn}(z_i)=1$ if $z_i>0$ and $=-1$ if  $z_i<0$, $(x)_+=\max(x, 0)$. To find the value of $\lambda$, we use the constraint $\lVert \bbx \rVert_1 = 1 $, which yields the following:
\begin{equation}
    \sum_{i=1}^p (|z_i| - \lambda)_+  = 1.
\end{equation}
To solve this equation for $\lambda$, we sort $|z_i|$, $i=1,\cdots, p$,  such that  $|z_{i_1}| > |z_{i_2}| > \cdots > |z_{i_p}|$.
We find $m$ such that the following inequalities are satisfied
\begin{equation}
  \begin{split}
    &|z_{i_j}|-\frac{1}{j-1}\Bigl(\sum_{k=1}^{j-1}|z_{i_k}|-1\Bigr)>0, j=2,\cdots, m\\
    &|z_{i_{m+1}}|-\frac{1}{m}\Bigl(\sum_{k=1}^{m}|z_{i_k}|-1\Bigr)\le 0. \\
  \end{split}
\end{equation}
After we find $m$, we have  $\sum_{j=1}^m (|z_{i_j}| - \lambda) = 1$, which yields 
$\lambda^* =  \frac{1}{m}(\sum_{j=1}^m |z_{i_j}| - 1)$. In summary, $\mathbb{P}_1(\bbz)$ can be written as
\begin{equation}
 \mathbb{P}_1(\bbz)=\begin{cases}
      \bbz, & \lVert \bbz \rVert_1 <1,\\
      \mathbb{S}(\bbz,\lambda^*), &\text{otherwise},
 \end{cases}
\end{equation}
where $\mathbb{S}(\bbz,\lambda^*)$ is a vector whose elements are $\mathbb{S}(z_i,\lambda^*)$, $i=1,\cdots, p$.

\section{Appendix D: Derivation of $\lambda_{\min}$ and $\lambda_{\max}$}
When $\lambda$ in \eqref{eq:cc} or \eqref{eq:ccprime} increases, the second term in the objective function  forces more cluster centers, $\bbu_i$'s, to be equal. When $\lambda\le  \lambda_{\min}$, all $\bbu_i$'s are unequal, whereas when $\lambda\ge  \lambda_{\max}$, all $\bbu_i$'s are equal. We will derive an estimate of $\lambda_{\min}$  and $\lambda_{\max}$ for the case where $\gamma=0$. 

 Let us consider an optimization problem that minimizes the following objective function:
\begin{equation} \label{eq:J1}
   J(\bbu)= \frac{1}{2}\Big(\lVert \bbx_i - \bbu_i \rVert_2^2 + \lVert \bbx_j - \bbu_j \rVert_2^2 \Big) + {\lambda} \omega_{ij}\lVert \bbu_{i} - \bbu_{j} \rVert_2,
\end{equation}
where  $\bbu = [\bbu_i^T, \bbu_j^T] \in \mathbb{R}^{2p \times 1}$. 
Let $\bbx = [\bbx_i^T, \bbx_j^T]^T \in \mathbb{R}^{2p \times 1}$, and $\bbD = [\bbI, -\bbI]$. Then, we can write \eqref{eq:J1} as
\begin{equation}
   J(\bbu)= \frac{1}{2}\lVert \bbx - \bbu \rVert_2^2 + {\lambda}\omega_{ij}\lVert\bbD\bbu\rVert_2.
\end{equation}
To minimize $J(\bbu)$, we formulate the following constrained optimization problem: 

\begin{equation}
\begin{split}
   &\min \frac{1}{2}\lVert \bbu - \bbx \rVert_2 + {\lambda}\omega_{ij}\lVert\bbv\rVert_2 \\
    &\text{subject to }\,\,\, \bbv = \bbD\bbu.
\end{split}  
\end{equation}

We will use the Lagrange dual function to solve this optimization problem, which will tell the condition under which $\bbu_i$ equals to $\bbu_j$  at the optimal solution.  The Lagrangian can be written as 
\begin{equation}\label{eq:lag}
    L(\bbu, \bbv, \bbeta) = \frac{1}{2}\lVert \bbu - \bbx \rVert_2^2 + {\lambda}\omega_{ij}\lVert\bbv\rVert_2 + \bbeta^T(\bbD\bbu - \bbv),
\end{equation}
where $\bbeta=[\eta_1, \cdots, \eta_p]^T$ and $\eta_i>0, \forall i$.  The gradient of  $L(\bbu, \bbv, \nu)$  w.r.t $\bbu$ is $\nabla_{\bbu}L =  \bbu - \bbx + \bbD^T\bbeta$.  Letting $\nabla_{\bbu}L=0$, we obtain
\begin{equation}\label{eq:u}
    \bbu = \bbx - \bbD^T\bbeta.
\end{equation}
Substituting $\bbu$ into \eqref{eq:lag} and using the identity $\bbD\bbD^T=2\bbI$,  $L(\bbu, \bbv, \bbeta)$ becomes 
\begin{equation}
    L(\bbv, \bbeta)=-\lVert\bbeta\rVert_2^2 +  \bbeta^T\bbD \bbx + {\lambda}\omega_{ij}\lVert\bbv\rVert_2  - \bbeta^T\bbv.
\end{equation}
To minimize   $L(\bbv, \bbeta)$ w.r.t $\bbv$, we need to minimize $l(\bbv)={\lambda}\omega_{ij}\lVert\bbv\rVert_2  - \bbeta^T\bbv$. Let us write $\bbv=c\bbz$, where $c=\lVert \bbv\rVert_2 \ge 0$, and $\lVert \bbz\rVert_2=1$.  Then, we have $l(\bbv)=c(\lambda\omega_{ij}-\bbeta^t\bbz)$. Minimizing $l(\bbv)$ w.r.t $\bbz$, we get $l(\bbv)=c(\lambda\omega_{ij}-\lVert\bbeta^t\rVert_2)$. Apparently, when $\lVert\bbeta\rVert_2\le \lambda\omega_{ij}$, the minimum of $l(\bbv)$ is zero with $c=0$; otherwise, $l(\bbv)$ has no minimum. 

Therefore, the Lagrange dual function is 
\begin{equation}
    L(\bbeta)=-\lVert\bbeta\rVert_2^2 +  \bbeta^T\bbD \bbx,
\end{equation}
and the Lagrange dual problem is 
\begin{equation}
\begin{split}
    \max_{\bbeta}\{ -\lVert\bbeta\rVert_2^2 +  \bbeta^T\bbD \bbx \}= \min_{\bbeta} \lVert \bbeta - \frac{\bbD \bbx}{2} \rVert_2^2 \\
    \text{s.t. } \lVert\bbeta\rVert_2\le \lambda\omega_{ij} \text{~and~} \eta_i>0, i=1,\cdots, p.
\end{split}
\end{equation}
The optimal solution of the dual problem is 
\begin{equation}
    \bbeta^* = \begin{cases}
    \frac{1}{2}\bbD\,\bbx \,,\, &\lVert\frac{\bbD\,\bbx}{2}\rVert_2 \leq {\lambda\,\omega_{ij}} \\
     \frac{\lambda\,\omega_{ij}}{\lVert\bbD\bbx\rVert_2}\bbD\bbx   &\text{otherwise}
    \end{cases}.
\end{equation}
Substituting $\bbeta^*$ into \eqref{eq:u}, we see that the optimal solution of the primal problem, $\bbu_i^*$ and $\bbu_j^*$, has the following properties: $\bbu_i^*=\bbu_j^*=\frac{1}{2}(\bbx_i+\bbx_j)$ when $\lVert\frac{\bbD\,\bbx}{2}\rVert_2 \leq {\lambda\,\omega_{ij}}$, and $\bbu_i^*\ne\bbu_j^*$, otherwise.

In other words, if $\lambda<  \lVert \bbx_i - \bbx_j \rVert_2/(2\omega_{ij})$, $\bbx_i$ and $\bbx_j$ are not in the same cluster, since  $\bbu_i^*\ne\bbu_j^*$.  To ensure any pair of data points are not in the sample cluster, we get $\lambda_{\min}$ as follows
\begin{equation}
     \lambda_{\min} < \min_{(i,j)\in\ccalE}\frac{\lVert \bbx_i - \bbx_j \rVert_2}{2\omega_{ij}}.
\end{equation}
Similarly, to make all data points to be in the same cluster, we get the $\lambda_{\max}$ as follows
\begin{equation}
     \lambda_{\max} \ge \max_{(i,j)\in\ccalE}\frac{\lVert \bbx_i - \bbx_j \rVert_2}{2\omega_{ij}}.
\end{equation}

\section{Appendix E: Derivation of $\gamma_{\max}$}
The subgradient of $\tdf_{\mu}(\bbU)$ in \eqref{eq:ccprox} can be written as 
\begin{equation}\label{eq:tdfsubg}
    \nabla \tdf_{\mu}(\bbU)  = \bbX - \bbU + \lambda \bbA \bbC^T + \gamma \sum_{k=1}^p \nu_{k} \frac{\partial \lVert \bba_k \rVert_2}{\partial \bbU},
\end{equation}
where  $\frac{\partial \lVert \bba_k \rVert_2}{\partial \bbU}$ stands for the subgradient of $\partial \lVert \bba_k \rVert_2$ w.r.t. $\bbU$. Recall that $\bba_k$ is the $k$ the row of $\bbU$. Therefore,  we have $\frac{\partial \lVert \bba_k \rVert_2}{\partial \bbU} = [\bb0, ..., \frac{\partial \lVert \bba_k \rVert_2}{\partial \bba_k^T}, ..., \bb0]^T$. Let us denote the $k$th row of a matrix $\bbX$ as $\bbX_{k,\cdot}$. Then, \eqref{eq:tdfsubg} can be written as
\begin{equation}\label{eq:tdfsubg2}
    \nabla \tdf_{\mu}(\bbU)_{k,\cdot} = \bbX_{k,\cdot} - \bbU_{k,\cdot} + \lambda(\bbA \bbC^T)_{k,\cdot} + \gamma \nu_{k}\bbs_k ,\, k = 1, \cdots ,p,
\end{equation}
where $\bbs_k=\frac{\partial \lVert \bba_k \rVert_2}{\partial \bba_k}=\bba_k^T /  \lVert \bba_k \rVert_2$ if $\bba_k \ne \bb0$, or $\bbs_k$ is a vector with $\lVert \bbs_k \rVert_2 \leq 1$ if $\bba_k = \bb0$. 

When $\gamma\ge \gamma_{\max}$, all $\bba_k$ (equivalently $\bbU_{k,\cdot}$), $k=1,\cdots, p$, are shrunk to zero, which results in $\bbA=0$. Substituting $\bbU=0$ and $\bbA=0$ into \eqref{eq:tdfsubg2} and letting $\nabla \tdf_{\mu}(\bbU)_{k,\cdot}=0$, we obtain
\begin{equation}
    \bbX_{k,.} = -\gamma_{\max} \nu_{k} \bbs_k,\, k = 1,\cdots,p.
\end{equation}
Since  $\lVert \bbs_k \rVert_2 \leq 1$, we have
$\lVert \bbX_{k,.} \rVert_2 \leq \gamma_{\max} \nu_{k}$, $k=1,\cdots, p$, which can be summarized as
\begin{equation}
    \gamma_{\max} = \max_{k=1,...,p} \frac{\lVert \bbX_{k,.} \rVert_2}{\nu_k}.
\end{equation}


\end{document}